# Variational Dual-Tree Framework for Large-Scale Transition Matrix Approximation

**Saeed Amizadeh**
Intelligent Systems Program
University of Pittsburgh
Pittsburgh, PA 15213

**Bo Thiesson**
Microsoft Research
Redmond, WA 98052

**Milos Hauskrecht**
Department of Computer Science
University of Pittsburgh
Pittsburgh, PA 15213

## Abstract

In recent years, non-parametric methods utilizing random walks on graphs have been used to solve a wide range of machine learning problems, but in their simplest form they do not scale well due to the quadratic complexity. In this paper, a new dual-tree based variational approach for approximating the transition matrix and efficiently performing the random walk is proposed. The approach exploits a connection between kernel density estimation, mixture modeling, and random walk on graphs in an optimization of the transition matrix for the data graph that ties together edge transitions probabilities that are similar. Compared to the de facto standard approximation method based on k-nearest-neighbors, we demonstrate order of magnitudes speedup without sacrificing accuracy for Label Propagation tasks on benchmark data sets in semi-supervised learning.

## 1 Introduction

Non-parametric methods utilizing random-walks on graphs have become very popular in Machine Learning during the last decade. These methods have been applied to solve Machine Learning problems as diverse as clustering (von Luxburg, 2007), dimensionality reduction (Lafon and Lee, 2006), semi-supervised learning (SSL) (Zhu, 2005), supervised learning (Yu et al., 2005), link analysis (Ng et al., 2001), and others. Unfortunately, these methods suffer from one fundamental and recurring problem: the quadratic dependency on the number of examples, which subsequently affects their scalability and applicability to large-scale data sets. Many different approximation frameworks have been proposed in the literature to tackle the above problem. Typically, these methods work by *sparsifying* the underlying graph representation by either reducing the number of nodes representing data points or reducing the number of edges representing data points similarities.

The first class of techniques aims to reduce the number of nodes in the graph from $N$ to $M$ ($M \ll N$). The reduction is carried out in various ways: random sampling (Kumar et al., 2009; Amizadeh et al., 2011), mixture modeling (Zhu and Lafferty, 2005), non-negative matrix factorization (Yu et al., 2005), sparse griding (Garcke and Griebel, 2005), Latent Markov Analysis (Lin, 2003), etc. A common problem with these techniques is that it is not clear how fast one should increase $M$ when $N$ increases. Subsequently, although the memory and the matrix-vector multiplication complexities are reduced to $O(M^2)$, we go back to $O(N^2)$ in effect if $M$ changes with the same rate as $N$ does.

The second class of techniques tries to sparsify the edges in the graph. $k$-nearest-neighbor graphs (Zhu, 2005) and $b$-matching (Jebara et al., 2009) are among the most famous methods in this group. The main concern with these methods, however, is the computational complexity of building these sparse graphs. Even with the smart speed-up techniques, such as $k$-nearest-neighbor graphs (Liaw et al., 2010; Moore, 1991), the actual time for building the full sparse graph can vary in practice.

A third class of methods goes around the problem of finding a sparse graph representation by trying to approximate the quantity of interest in the problem directly. Manifold regularization techniques (Belkin et al., 2006; Tsang and Kwok, 2006) directly find a smooth function on the graph. Fergus et al. (2009); Nadler et al. (2006); Amizadeh et al. (2012) try to directly estimate the eigen decomposition of the underlying transition matrix in the limit assuming factorized underlying distribution. Although these methods can be very effective, they are also task specific and do not give us an explicit approximation of the graph similarity (transition) structure.

The framework proposed in this paper directly approximates the transition matrix of the data graph. Our method is similar to the second group of approximation techniques in the sense that it does not reduce the graph nodes. However, instead of zeroing out the edges (like in the second group), our framework groups the edges together to *share* edge transition probabilities that are similar. To do so, we use a recent advancement from Thiesson and Kim (2012), which for a very different task (fast mode-seeking) developed a fast dual-tree based (Gray and Moore, 2000) variational framework that exploits a mixture modeling view on kernel density estimation. Mixture modeling and the random walk on graphs are also closely related (Yu et al., 2005). Based on this connection between kernel density estimation, mixture modeling, and random walk on graphs, we extend the framework in Thiesson and Kim (2012) to a fast and scalable framework for 1) transition matrix approximation and 2) inference by the random walk.

By further exploiting the connection between mixture modeling and random walk, we also propose a lower-bound log-likelihood optimization technique to find the optimal bandwidth for the Gaussian similarity kernel (which is a very popular kernel for constructing transition matrices on data graphs). Moreover, using our framework, one can adjust the trade-off between accuracy and efficiency by *refining* the model. As we show in the paper, at its coarsest level of refinement, our framework achieves complexity order of $O(N^{1.5} \log N)$ for construction, $O(N)$ inference via matrix-vector multiplication, and memory consumption of $O(N)$, which improves the performance of graph sparsification methods reviewed earlier. To demonstrate the speed-up in practice, we experimentally show that without compromising much in terms of accuracy, our framework can build, represent and operate on random-walk transition matrices orders of magnitude faster than the baseline methods.

## 2 Background

Kernel density estimation plays an important conceptual role in this paper. Let $\mathcal{M} = \{m_1, m_2, \ldots, m_N\}$ denote a set of kernel centers corresponding to observed data $\mathcal{D}$ in $\mathbb{R}^d$. The kernel density estimate for any data point $x \in \mathbb{R}^d$ is then defined by the following weighted sum or mixture model:

$$p(x) = \sum_j p(m_j) p(x|m_j) \quad (1)$$

where $p(m_j) = \frac{1}{N}$, $p(x|m_j) = \frac{1}{Z_\sigma} k(x, m_j; \sigma)$, and $k$ is a kernel profile with bandwidth $\sigma$ and normalization constant $Z_\sigma$. For the commonly used Gaussian kernel, $k(x, m_j; \sigma) = \exp(-\|x - m_j\|^2 / 2\sigma^2)$ and $Z_\sigma = (2\pi\sigma^2)^{\frac{d}{2}}$.

Now let us assume we want kernel density estimates for data points $\mathcal{D} = \{x_1, x_2, \ldots, x_N\}$ in $\mathbb{R}^d$ that also define the kernel centers. That is, $\mathcal{D} = \mathcal{M}$, and we use the $\mathcal{D}$ and $\mathcal{M}$ notation to emphasize the two different roles every example takes - one as a data point, the other as a kernel center. Let us exclude the kernel centered at $m_i = x_i$ from the kernel density estimate at that data point. In this case, the estimate at $x_i \in \mathcal{D}$ can be defined as the $N-1$ component mixture model:

$$p(x_i) = \sum_{j \neq i} p(m_j) p(x_i|m_j) = \sum_{j \neq i} \frac{1}{N-1} \frac{1}{Z_\sigma} k(x_i, m_j; \sigma), \quad (2)$$

where $p(m_j) = 1/(N-1)$.

In the mixture model interpretation for the kernel density estimate, the posterior kernel membership probability for a data point can be expressed as

$$p(m_j|x_i) = \frac{p(m_j) p(x_i|m_j)}{p(x_i)} = \frac{k(x_i, m_j; \sigma)}{\sum_{l=1}^{N} k(x_i, m_l; \sigma)} \quad (3)$$

These posteriors form a matrix $P = [p_{ij}]_{N \times N}$ where $p_{ij} = p(m_j|x_i)$ for $i \neq j$ are the true posteriors, and $p_{ij} = 0$ for $i = j$ are neutral elements added for later notational convenience. Note that in this representation, each row in $P$ holds a posterior distribution $p_{i\cdot} \triangleq \{p_{ij}|j = 1, \ldots, N\}$.

The complexity of computing and representing $P$ is $O(N^2)$, which is costly for large-scale data sets with large $N$. In this work, we seek ways of alleviating the problem by taking the advantage of the cluster structure in data (and kernels) that would let us *share* posteriors among groups of data points and kernels instead of computing them for each individual pair. More precisely, suppose that the data set $\mathcal{D}$ consists of two well-separated clusters in $\mathbb{R}^d$, denoted by $C_1$ and $C_2$. In this case, it is likely that the posteriors $p(m_j|x_i)$ are roughly similar for all $x_i \in C_1$ and $m_j \in C_2$ so that one can approximate all posteriors for the pairs in between these two clusters with one value. Notice how this approximation directly targets the posteriors and not the expensive intermediate kernel summation in the denominator of (3). This simple idea can be generalized recursively if we have nested cluster structure in the form of a *cluster hierarchy*. In this case, the computational savings can be dramatic.

## 3 Dual-tree Based Variational $P$

Whilst the approximation idea for $P$ outlined above may appear simple, it needs careful handcrafting in order to yield a practical and computationally efficient framework. We start building our framework by borrowing the ideas from the dual-tree-based variational approach proposed by Thiesson and Kim (2012).

## 3.1 Dual-tree Block Partitioning

The dual-tree-based variational approach in Thiesson and Kim (2012) maintains two tree structures; the *data partition tree* and the *kernel partition tree*, that hierarchically partition data points (kernels) into disjoint subsets, such that an intermediate node in a tree represents a subset of data points (or kernels) and leaves correspond into singleton sets. In this work we assume the structure of the two trees is identical, leading to the exactly same subsets of data points and kernels represented by the tree.

The main reason for introducing the partition tree is to define relations and permit inferences for groups of related data points and kernels without the need to treat them individually. More specifically, we use the partition tree to induce a *block partition* of the matrix $P$, where all posteriors within the block are forced to be equal. We also refer to this partition as *block constraints* on $P$. Formally, a *valid* block partition $\mathcal{B}$ defines a mutually exclusive and exhaustive partition of $P$ into blocks (or sub-matrices) $(A, B) \in \mathcal{B}$, where $A$ and $B$ are two *non-overlapping* subtrees in the partition tree. That is, $A \cap B = \emptyset$, in the sense that data-leaves in the subtree under $A$ and kernel-leaves in the subtree under $B$ do not overlap. (To maintain the convenient matrix representation of $P$, the singleton blocks representing the neutral diagonal elements in $P$ are added to this partition.) Figure 1a) shows a small example with a valid block partition for a partition tree built for six data points (kernels). This block partition will, for example, enforce the block constraint $p_{13} = p_{14} = p_{23} = p_{24}$ for the block $(A, B) = (1{-}2, 3{-}4)$ (where $a{-}b$ denotes $a$ through $b$).

To represent the block-constrained $P$ matrix more compactly, we utilize the so-called *marked partition tree (MPT)* that annotates the partition tree by explicitly linking data groups to kernel groups in the block partition: for each block $(A, B) \in \mathcal{B}$, the data-node $A$ is marked with the matching kernel-node $B$. Each node $A$ in the MPT will therefore contain a, possibly empty, list of marked $B$ nodes. We will denote this list of marks as $A_{mkd} \triangleq \{B | (A, B) \in \mathcal{B}\}$. Figure 1b) shows the MPT corresponding to the partition in Figure 1a). For example, the mark of kernels $B = 3{-}4$ at the node representing the data $A = 1{-}2$ corresponds to the same block constraint $(A, B) = (1{-}2, 3{-}4)$, as mentioned above. It is the only mark for this data node, and therefore $A_{mkd} = 1{-}2_{mkd} = \{3{-}4\}$. As another example, the list of marks for node $A = 5$ has two elements; $A_{mkd} = 5_{mkd} = \{5, 6\}$. An important technical observation, that we will use in the next section, is that each path from a leaf to the root in the MPT corresponds to the row indexed by that leaf in the block partition matrix for $P$. By storing the posterior probabilities at the marks in the MPT, we can therefore extract the entire posterior distribution $p_{i\cdot}$ by starting at the leaf that represents the data point $x_i$ and follow the path to the root.

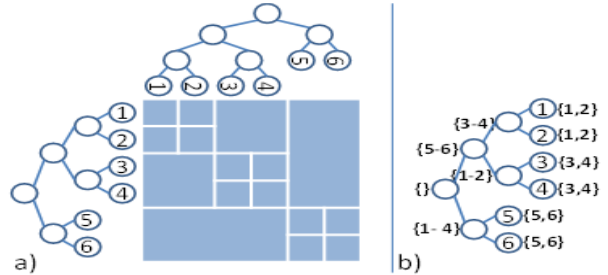

Figure 1: a) A block partition (table) for data partition tree (left) and identical kernel partition tree (top). b) MPT representation of the block partition.

Clearly, there are more than one valid partitions of $P$; in fact, any further *refinement* of a valid partition results in another valid partition with increased number of blocks. We postpone the discussion of how to choose an initial valid partition and how to refine it to Section 4.4. For now, let us assume that we are given a valid partition $\mathcal{B}$ of $P$ with $|\mathcal{B}|$ number of blocks. By insisting on the block constraint of equality for all posteriors in a given block, the number of parameters in $P$ is effectively reduced from $N^2$ to $|\mathcal{B}|$.

## 3.2 Optimization of Block Partitioning

In general, the block partition approach links groups of data points and kernels together such that any specific link is no longer able to distinguish its components. The key question now is whether we can cast the block partitioning problem into an optimization framework that can approximate well the true unconstrained $P$. We address the question by using a variational approximation approach. In this case, each block $(A, B) \in \mathcal{B}$ is assigned a single variational parameter $q_{AB}$ that approximates the posteriors $p_{ij}$ for all $x_i \in A$ and $m_j \in B$ in that block. That is, for $Q = [q_{ij}]_{N \times N}$

$$q_{ij} = q_{AB} \text{ s.t. } (A, B) \in \mathcal{B}, x_i \in A, m_j \in B \quad (4)$$

Recall that in our setup, $q_{ij} = p_{ij} = 0$ for $i = j$.

The trick to solve the problem rests on expressing a variational lower bound for the log-likelihood of data:

$$\log p(\mathcal{D}) = \sum_i \log p(x_i) = \sum_i \log \sum_{j \neq i} p(m_j) p(x_i \mid m_j)$$
$$= \sum_i \log \sum_{j \neq i} \frac{q_{ij}}{q_{ij}} p(m_j) p(x_i \mid m_j)$$
$$\geq \sum_i \sum_{j \neq i} q_{ij} \log \frac{p(m_j) p(x_i \mid m_j)}{q_{ij}} \quad (5)$$

$$= \log p(\mathcal{D}) - \sum_i D_{KL}(q_{i\cdot}\|p_{i\cdot}) \triangleq \ell(\mathcal{D}), \quad (6)$$

where $D_{KL}(\cdot\|\cdot)$ is the KL-divergence between two distributions. Notice that optimizing the lower bound $\ell(\mathcal{D})$ with respect to $Q$ corresponds to minimizing the KL-divergence between the two distributions $q_{i\cdot}$ and $p_{i\cdot}$ for all $i$, since this is the only term in (6) that depends on $Q$. The block-constrained $Q$ matrix is therefore a good approximation for the unconstrained $P$.

Let us now explicitly insert the block constraints from (4) into the expression for $\ell(\mathcal{D})$ in (5). With $p(m_i) = 1/(N-1)$ and the Gaussian (normalized) kernel for $p(x_i|m_j)$, the optimization reduces to finding the variational parameters $q_{AB}$ that maximizes

$$\ell(\mathcal{D}) = c - \frac{1}{2\sigma^2} \sum_{(A,B)\in\mathcal{B}} q_{AB} \cdot D^2_{AB}$$
$$- \sum_{(A,B)\in\mathcal{B}} |A||B| \cdot q_{AB} \log q_{AB}, \quad (7)$$

where

$$c = -N \log\left((2\pi)^{d/2}\sigma^d(N-1)\right)$$
$$D^2_{AB} = \sum_{x_i\in A}\sum_{m_j\in B} \|x_i - m_j\|^2. \quad (8)$$

Thiesson and Kim (2012)[Algorithm 3] has developed a recursive algorithm that solves this optimization for all $q_{AB}$, $(A,B)\in\mathcal{B}$ in $O(|\mathcal{B}|)$ time. Specifically, their solution avoids the quadratic complexity that a direct computation of the Euclidean distances in (8) would demand by factorizing $D^2_{AB}$ into data-specific and kernel-specific statistics as follows

$$D^2_{AB} = |A|S_2(B) + |B|S_2(A) - 2S_1(A)^T S_1(B), \quad (9)$$

where $S_1(A) = \sum_{x\in A} x$ and $S_2(A) = \sum_{x\in A} x^T x$ are the statistics of subtree $A$. These statistics can be incrementally computed and stored while the shared partition tree is being built; an $O(N)$ computation. Using these statistics, $D^2_{AB}$ is computed in $O(1)$.

Now, the question is how to efficiently construct the shared partition tree from data. Well-known, efficient partition-tree construction methods include *anchor* tree (Moore, 2000), *kd*-tree (Moore, 1991) and *cover* tree (Ram et al., 2009; Beygelzimer et al., 2006) construction. We have used the anchor tree. With a relatively balanced tree, it takes $O(N^{1.5}\log N + |\mathcal{B}|)$ time and $O(|\mathcal{B}|)$ memory in total to build and store the variational approximation for the posterior matrix $P$ from data. See the supplementary Appendix for more details on the anchor tree construction complexity.

## 4 Application to Random Walk

In this section, we show how the variational mixture modeling framework can be utilized to tackle large-scale random walk problems. The random walk here is defined on the graph $\mathcal{G}$ which is an undirected similarity graph whose nodes are the data points in $\mathcal{D}$ and the edge between nodes $x_i$ and $x_j$ is assigned the similarity weight $s_{ij} = \exp(-\|x_i - x_j\|^2/2\sigma^2)$.

### 4.1 Variational Random Walk

As we saw in Section 2, $p(m_j|x_i)$ models the posterior membership of data point $x_i$ to the Gaussian kernel $m_j$. However, $p(m_j|x_i)$ can be interpreted from a completely different view as the probability of jumping from data point $x_i$ to data point $m_j$ in a random walk on $\mathcal{G}$. In fact, (3) is the same formula that is used to compute the transition probabilities on data graphs, when the Gaussian similarity is used (Lafon and Lee, 2006). As a result, the matrix $P$ from the previous section can be seen as the transition probability matrix for a random walk on $\mathcal{G}$. Subsequently, the corresponding variational approximation of $Q$ is in fact the approximation of the transition probability matrix using only $|\mathcal{B}|$ number of parameters (blocks); in other words, $q_{AB}$ approximates the probability of jumping from a data point in $A$ to a data point in $B$. This is, in particular, important because it enables us to compute and store the transition probability matrix in $O(N^{1.5}\log N + |\mathcal{B}|)$ time and $O(|\mathcal{B}|)$ memory for large-scale problems.

In the light of this random walk view on $Q$, the lower bound log-likelihood $\ell(\mathcal{D})$ in (7) will also have a new interpretation. More precisely, the second term in (7) can be reformulated as:

$$-\frac{1}{2\sigma^2}\sum_{(A,B)\in\mathcal{B}} q_{AB}\cdot D^2_{AB} = -\frac{1}{2\sigma^2}\sum_{i,j} q_{ij}\cdot \bar{d}_{ij} \quad (10)$$

where, $q_{ij}$ is defined as in (4) and $\bar{d}_{ij} = D^2_{AB}/|A||B|$ is the *block-average distance* such that $(A,B)\in\mathcal{B}$, $x_i \in A$ and $m_j \in B$. This is, in fact, a common optimization term in similarity-graph learning from mutual distances (Jebara et al., 2009) and can be more compactly represented as $-\frac{1}{2\sigma}tr(Q\bar{D})$, where $Q = [q_{ij}]_{N\times N}$ and $\bar{D} = [\bar{d}_{ij}]_{N\times N}$ are the similarity and distance matrices, respectively. However, there is one problem with a maximization of this term: it will make each point connect to its closest neighbor with $q_{ij} = 1$ (and $q_{ij} = 0$ for the rest). In other words, the similarity graph will be highly disconnected. This is where the third term in (7) benefits the new interpretation. One can rewrite this term as:

$$-\sum_{(A,B)\in\mathcal{B}} |A||B|\cdot q_{AB}\log q_{AB} = -\sum_{i,j} q_{ij}\log q_{ij}$$
$$= \sum_{i=1}^N H(q_{i\cdot}) \quad (11)$$

where $H(q_{i.})$ is the entropy of the transition probability distribution from data point $x_i$. As opposed to (10), the term in (11) is maximized by a uniform distribution over the outgoing probabilities at each data point $x_i$; that is, a fully connected graph with equal transition probabilities. The third term in (7) therefore acts as the *regularizer* in the learning of the similarity-graph, trying to keep it connected. The trade-off between the second and the third terms is adjusted by the coefficient $1/2\sigma^2$: increasing $\sigma$ will leave the graph more connected.

### 4.2 Learning $\sigma$

Finding the transition probabilities for a random walk by the variational optimization of (7) has a side advantage too: (7) is a *quasi-concave* function of the bandwidth $\sigma$, which means that, given $q_{AB}$'s fixed, one can find the optimal bandwidth that maximizes the log-likelihood lower bound. By taking the derivative and solving for $\sigma$, the closed form solution is:

$$\sigma^* = \sqrt{\frac{\sum_{(A,B)\in\mathcal{B}} q_{AB} \cdot D_{AB}^2}{Nd}} \quad (12)$$

In the special case where each element of matrix $P$ is a singleton block (i.e. the most refined case), one can find $\sigma^*$ independent of $q_{AB}$'s values. To do so, we first form the following log-likelihood lower bound using the Jensen inequality:

$$\log p(\mathcal{D}) \geq \sum_i \sum_{j \neq i} p(m_j) \log p(x_i \mid m_j) \quad (13)$$

By maximizing the right-hand side w.r.t. $\sigma$, we get:

$$\sigma^* = \frac{1}{N}\sqrt{\frac{\sum_i \sum_{j\neq i} \|x_i - x_j\|^2}{d}} \quad (14)$$

In general, due to the dependence of $\sigma^*$ to $q_{AB}$'s, we alternate the optimization of $q_{AB}$'s and $\sigma$ in our framework. In practice, we have observed that the convergence of this alternate optimization is fast and not sensitive to the initial value of $\sigma$.

### 4.3 Fast Inference

In the previous subsections, we saw how the variational dual-tree based framework can be used to efficiently build and store a variational transition matrix $Q$ of a random walk. We will now demonstrate that the block structure of this transition matrix can be very useful for further inference in similarity-graph based learning algorithms. In particular, using the MPT representation of $Q$, we can efficiently compute the multiplication with an arbitrary vector $Y = (y_1, y_2, \ldots, y_n)^T$ of observations to achieve $\hat{Y} = QY \simeq PY$ in $O(|\mathcal{B}|)$ rather than $O(N^2)$ computations.

---

**Algorithm 1** Calculate $\hat{Y} = QY$

**Input**: MPT with $\{q_{AB} : B \in A_{mkd}\}$ on each node $A$, and $(x_i, y_i)$ on each leaf.
**Output**: $\hat{y}_i$ on each leaf.

$A = \textsc{Root}(\text{MPT})$
$\textsc{CollectUp}(A)$
$\textsc{DistributeDown}(A, 0)$

**function** $\textsc{CollectUp}(A)$
  **if** $\textsc{IsLeaf}(A)$ **then**
    $T_A = y_A$     //null-vector if $y_A$ is not observed
  **else**
    $\textsc{CollectUp}(A_l)$
    $\textsc{CollectUp}(A_r)$
    $T_A = T_{A_l} + T_{A_r}$
  **end if**
**end function**

**function** $\textsc{DistributeDown}(A, py)$
  **for all** marks $B \in A_{mkd}$ **do**
    $py \mathrel{+}= |B|q_{AB}T_A$
  **end for**
  **if** $\textsc{IsLeaf}(A)$ **then**
    $\hat{y}_A = py$
  **else**
    $\textsc{DistributeDown}(A_l, py)$
    $\textsc{DistributeDown}(A_r, py)$
  **end if**
**end function**

---

Algorithm 1 describes the $O(|\mathcal{B}|)$ computation of $\hat{Y}$. The algorithm assumes that each $y_i$ is stored at the corresponding leaf $x_i$ in the MPT–e.g., by association prior to constructing the MPT. Alternatively, an index to the leaves can be constructed in $O(N)$ time. The algorithm starts with a CollectUp phase that traverses the MPT bottom-up and stores incrementally computed sum-statistics at each node $A$, as

$$T_A = \sum_{x_i \in A} y_i = T_{A_l} + T_{A_r},$$

where $A_l$, $A_r$ are the left and right children of $A$. This is an $O(N)$ computation. Let $\mathcal{B}(x_i) \triangleq \{(A, B) \in \mathcal{B} \mid x_i \in A\}$ denote the set of marked blocks that can be experienced on the path in the MPT from leaf $x_i$ to the root. For example, in Figure 1, $\mathcal{B}(3) = \{(3,3), (3,4), (3\text{–}4, 1\text{–}2), (1\text{–}4, 5\text{–}6)\}$. With each $q_{AB}$ stored at the node $A$ marked with $B$ in the MPT, a DistributeDown phase now conceptually computes

$$\hat{y}_i = \sum_{(A,B) \in \mathcal{B}(x_i)} |B| q_{AB} T_A \simeq \sum_j p_{ij} y_j,$$

by following the path from each leaf to the root in the MPT. The more efficient implementation in Algorithm 1 traverses the MPT top-down, avoiding re-calculations of the shared terms for the updates by propagating the value of the shared terms through the variable $py$. This traversal has complexity $O(|\mathcal{B}|)$. On completion of the algorithm the leaves in the MPT will contain $\hat{Y} = \{y_i\}_{i=1}^N$. The fast matrix-vector multiplication is a significant achievement because it allows us to significantly speed up any algorithm with this computational bottleneck. Two such algorithms are Label Propagation (LP) (Zhou et al., 2003) used for Semi-supervised Learning and Link Analysis (see, e.g., Ng et al. (2001)), and Arnoldi Iteration (Saad, 1992) used for spectral decomposition. More specifically, given a similarity graph over the data points $x_i$, the LP algorithm iteratively updates the label matrix $Y = [y_{ij}]_{N \times C}$ (where $C$ is the number of label classes) at time $t+1$ by propagating the labels at time $t$ one step forward according to the transition matrix $P$:

$$Y^{(t+1)} \leftarrow \alpha P Y^{(t)} + (1-\alpha) Y^0 \quad (15)$$

Here, the matrix $Y^0$ encodes the initial labeling of data such that $y_{ij}^0 = 1$ for $\text{label}(x_i) = y_i = j$, and $y_{ij}^0 = 0$ otherwise. The coefficient $\alpha \in (0,1)$ determines how fast the label values are updated. The repeated matrix multiplication in (15) definitely poses a bottleneck for large-scale problems, and this is where our fast framework comes into play.

### 4.4 Partitioning and Refinement

So far, we have assumed that a valid partition $\mathcal{B}$ of $P$ with $|\mathcal{B}|$ number of blocks is given. In this section, we illustrate how to construct an initial valid partition of $P$ and how to further refine it to increase accuracy of the model. We should note that the methods for partitioning and refinement in this section are different from the ones in Thiesson and Kim (2012).

Let $\mathcal{B}_{diag}$ denote the $N$ neutral singleton blocks that appear on the diagonal of $P$. The coarsest (with the smallest number of blocks) valid partition $\mathcal{B}_c$ for $P$ is achieved when for every block $(A,B) \in \mathcal{B}_c = \mathcal{B} \setminus \mathcal{B}_{diag}$, we have that $A$ and $B$ are sibling subtrees in the partition tree. (Recall that data and kernels are partitioned by the same tree.) Any other partition will either not be a valid partition conforming with the partition tree, or it will have a larger number of blocks. The number of blocks in $\mathcal{B}_c$, therefore equals twice the number of inner nodes in the anchor tree; i.e. $|\mathcal{B}_c| = 2(N-1)$. Figure 1 is an example of a coarsest valid block partition. On the other hand, the most refined partition is achieved when $\mathcal{B}_r = \mathcal{B} \setminus \mathcal{B}_{diag}$ contains $N^2 - N$ singleton blocks. Hence, the number of blocks in a valid partition $|\mathcal{B}|$ can vary between $O(N)$ and $O(N^2)$. As we saw in the previous sections, $|\mathcal{B}|$ plays a crucial role in the computational performance of the whole framework and we therefore want to keep it as small as possible. On the other hand, keeping $|\mathcal{B}|$ too small may excessively compromise the accuracy of the model. Therefore, the rational approach would be to start with the coarsest partition $\mathcal{B}_c$ and split the blocks in $\mathcal{B}_c$ into smaller ones only if needed. This process is called *refinement*. As we refine more blocks, the accuracy of the model effectively increases while its computational performance degrades. Note that a block $(A,B)$ can be refined in two ways: either *vertically* into $\{(A_l, B), (A_r, B)\}$ or *horizontally* into $\{(A, B_l), (A, B_r)\}$. Here the subscript $r$ ($l$) denotes the right (left) child node.

After any refinement, we can re-optimize (7) to find the new variational approximation $Q$ for $P$. Importantly, any refinement loosens the constraints imposed on $Q$, implying that the KL-divergence from $P$ cannot increase. From (6) we can therefore easily see that a refinement is likely to increase the log-likelihood lower bound $\ell(\mathcal{D})$, and can never decrease it. Intuitively, we want to refine those blocks which increase $\ell(\mathcal{D})$ the most. To find such blocks, one need to refine each block in each direction (i.e. horizonal and vertical) one at a time, re-optimize $Q$, find the difference between the new $\ell(\mathcal{D})$ and the old one (aka *log-likelihood gain*), and finally pick the refinements with maximum difference. However, this is an expensive process since we need to perform re-optimization per each possible refinement. Now the question is, whether we can obtain an estimate of the log-likelihood gain for each possible refinement without performing re-optimization. The answer is positive for horizontal refinements, and rests on the fact that each row in $Q$ defines a (posterior) probability distribution and therefore must sum to one. In particular, this sum-to-one constraint can for $\mathcal{B}(x_i) \triangleq \{(A,B) \in \mathcal{B} \mid x_i \in A\}$ be expressed as

$$\sum_{(A,B) \in \mathcal{B}(x_i)} |B| \cdot q_{AB} = 1 \text{ for all } x_i \in \mathcal{D}. \quad (16)$$

Consider the horizontal refinement of $(A,B)$ into $\{(A, B_l), (A, B_r)\}$. By this refinement, in essence, we allow a random walk, where the probability of jumping from points in $A$ to the ones in $B_l$ (i.e. $q_{AB_l}$) is different from that of jumping from $A$ to $B_r$ (i.e. $q_{AB_r}$). If we keep the $q$ values for other blocks fixed, we can still *locally* change $q_{AB_l}$ and $q_{AB_r}$ to increase $\ell(\mathcal{D})$. Since the sum of the outgoing probabilities from each point is 1 (see (16)) and the other $q$'s are unchanged, the sum of the outgoing probabilities from $A$ to $B_l$ and $B_r$ must be equal to that of the old one from $A$ to $B$:

$$|B_l| q_{AB_l} + |B_r| q_{AB_r} = |B| q_{AB}. \quad (17)$$

Under this local constraint, we can find $q_{AB_l}$ and $q_{AB_r}$ in closed form such that $\ell(\mathcal{D})$ is maximized:

$$q_{AB_c} = \frac{|B| \exp(G_{AB_c}) q_{AB}}{\sum_{t \in \{l,r\}} |B_t| \exp(G_{AB_t})}, \ c \in \{l,r\} \quad (18)$$

where $G_{AB} = -D_{AB}^2 / (2\sigma^2 |A||B|)$. By inserting (18) in (7), we can compute the maximum log-likelihood gain for the horizontal refinement of $(A,B)$ as

$$\Delta_{AB}^h = \ell'(\mathcal{D}) - \ell(\mathcal{D})$$
$$= |A||B| q_{AB} \cdot \log\left(\frac{\sum_{t \in \{l,r\}} |B_t| \exp(G_{AB_t})}{|B| \exp(G_{AB})}\right), \quad (19)$$

where $\ell'(\mathcal{D})$ denotes the log-likelihood lower bound after the refinement. Note that the actual gain can be greater than $\Delta^h_{AB}$ because after a refinement, all $q$ values are re-optimized. In other words, $\Delta^h_{AB}$ is a lower bound for the actual gain and is only used to pick blocks for refinements. Unfortunately, for vertical refinements such a bound is not easily obtainable. The reason is that the constraints in (16) will not allow $q_{A_lB}$ and $q_{A_rB}$ to change if we fix the remaining $q$ values. The local optimization that we applied for the horizontal refinement can therefore not be applied to estimate a vertical refinement. Whenever we pick a block $(A, B)$ for vertical refinement, we therefore incorporate vertical refinements by applying the horizontal refinement to its symmetric counterpart $(B, A)$ if it also belongs to $\mathcal{B}$. We call this *symmetric refinement*.

Now by computing $\Delta^h_{AB}$ for all blocks, we greedily pick the block with the maximum gain and apply symmetric refinement. The newly created blocks are then added to the pool of blocks and the process is repeated until the number of blocks reaches the maximum allowable block number $|\mathcal{B}|_{max}$, which is an input argument to the algorithm. This greedy algorithm can be efficiently implemented using a priority queue.

## 5 Experiments

In this section, we present the experimental evaluation of our framework. In particular, we have evaluated how well our method performs for semi-supervised learning (SSL) using Label Propagation (LP) (Zhou et al., 2003). The LP algorithm starts with a partial vector of labels and iteratively *propagates* those labels over the underlying similarity graph to unlabeled nodes (see (15)). After $T$ iterations of propagation, the inferred labels at unlabeled nodes are compared against the true labels (which were held out) to compute Correct Classification Rate (CCR). We also measure the time (in ms) taken to build each model, propagate labels and refine each model. In our experiments, we set $T = 500$ and $\alpha = 0.01$. It should be noted that, here the goal is not to achieve a state-of-the-art SSL algorithm, but to relatively compare our framework to baselines in terms of efficiency and accuracy under the *same* conditions. This means that we have not tuned the SSL parameters to improve SSL; however, we use the same parameters for all competitor methods.

### 5.1 The Baselines

We have compared our method, *VariationalDT*, with two other methods for building and representing $P$. The first method is the straightforward computation of $P$ using (3). We refer to this representation as the *exact model*. In terms of computational complexity, it takes $O(N^2)$ to build, store, and multiply a vector to $P$ using the exact method. The second method is the *k-nearest-neighbor (kNN)* algorithm where each data point is connected only to its $k$ closest neighbors. In other words, the rows of matrix $P$ will contain only $k$ non-zero entries such that $P$ can be represented as a sparse matrix for small $k$'s. In that way, $k$NN zeros out many of these parameters to make $P$ sparse, as opposed to our VariationalDT method that groups and shares parameters in $P$. Note that we still assign weights to the $k$ edges for each data point using (3). This means that as we increase $k$ toward $N$, the model converges to the exact model. Therefore, $k$ acts as a tuning parameter to trade off between computational efficiency and accuracy (the same role that $|\mathcal{B}|$ plays in VariationalDT). We call $k$ and $|\mathcal{B}|$ the trade-off parameters.

Using the $k$NN representation, $P$ can be stored and multiplied by an arbitrary vector in $O(kN)$. For constructing a $k$NN graph directly, the computational complexity is $O(rN^2)$ where $r = \min\{k, \log N\}$. A smarter approach is to use a metric tree in order to avoid unnecessary distance computations. Moore (1991) proposed a speedup of the $k$NN graph construction that utilizes a *kd-tree*. In our implementation of $k$NN, we have used the same algorithm with the $kd$-tree replaced by the anchor tree, introduced in Section 3. We call this algorithm *fast kNN*. The computational analysis of fast $k$NN greatly depends on the distribution of data points in the space. In the best case, it takes $O(N(N^{0.5} \log N + k \log k))$ to build the $k$NN graph using fast $k$NN. However, in the worst case, the computational order is $O(N(N^{0.5} \log N + N \log k))$. Table 1 summarizes the computational complexity orders for the models compared in the experiments. Note that for $|\mathcal{B}| = kN$ for some integer $k$, the VariationalDT and $k$NN will have the same memory and multiplication complexity. To increase $k$ (or equivalently $|\mathcal{B}|$), one needs to refine the respective model. The last column in Table 1 contains the complexity of refining $k$ to $k+1$ (or equivalently $|\mathcal{B}|$ from $kN$ to $(k+1)N$) for $k$NN (or VariationalDT).

### 5.2 Experimental Setup

We have performed three experiments to evaluate the computational efficiency and accuracy of the models described above. It should be noted that for all methods, we have used the log-likelihood lower-bound technique in Section 4.2 to tune the bandwidth $\sigma$.

In the first experiment, we have run the LP algorithm for semi-supervised learning on the SecStr data set, one of the benchmark data sets for SSL (Chapelle et al., 2006). The data set consists of 83,679 different amino acids each of which is represented by 315 binary features. The task is to predict the secondary structure of the amino acids (2 classes). The goal in this experiment, as we increase the problem size $N$, is to study (a) the time needed to build the exact model, the coarsest Variational DT model (i.e. $|\mathcal{B}| = 2(N-1)$), and the coarsest $k$NN model (i.e. $k = 2$), (b) the time needed for multiplication in these models, and (c) CCR after label propagation given 10% labeled data. In particular, we draw samples of size $s$ from the data set and use each sample to construct a model. Once a model is built, we choose 10% of the sample randomly to be fed to the LP algorithm as the labeled partition. We repeat this process 5 times for each problem size $s$ and report the average result. Figure 2A) shows the construction time (in ms) for the three models as the problem size increases. As the plot shows, our method is orders of magnitude faster than the other two baselines (note that it is the log-log scale). In terms of the time needed for one multiplication, Figure 2B) shows that our method and $k$NN have similar complexity while both are order of magnitudes faster than the exact method. Note that Figure 2B) also shows the proportional memory usage for these methods because according to Table 1 multiplication and memory usage have the same complexity in all three methods. Finally, Figure 2C) depicts the CCR for the three models with different problem sizes. Although the exact model as expected is slightly more accurate than $k$NN and our method, the difference is not that big. In other words, by using VariationalDT we save orders of magnitudes in memory and CPU (i.e. both the construction and multiplication times) while compromising a little on accuracy.

| Models | Construction | Memory | Multiplication | Refinement |
|---|---|---|---|---|
| **Exact** | $O(N^2)$ | $O(N^2)$ | $O(N^2)$ | N/A |
| **Fast $k$NN** | $O\big(N(N^{0.5}\log N + h\log k)\big)^*$ | $O(kN)$ | $O(kN)$ | $O\big(N(\log N + N\log k)\big)$ |
| **Variational DT** | $O(N^{1.5}\log N + |\mathcal{B}|)$ | $O(|\mathcal{B}|)$ | $O(|\mathcal{B}|)$ | $O(|\mathcal{B}|\log|\mathcal{B}|)$ |

Table 1: Theoretical complexity analysis results. (*) $h$ is equal to $k$ in the best case and $N$ in the worst case.

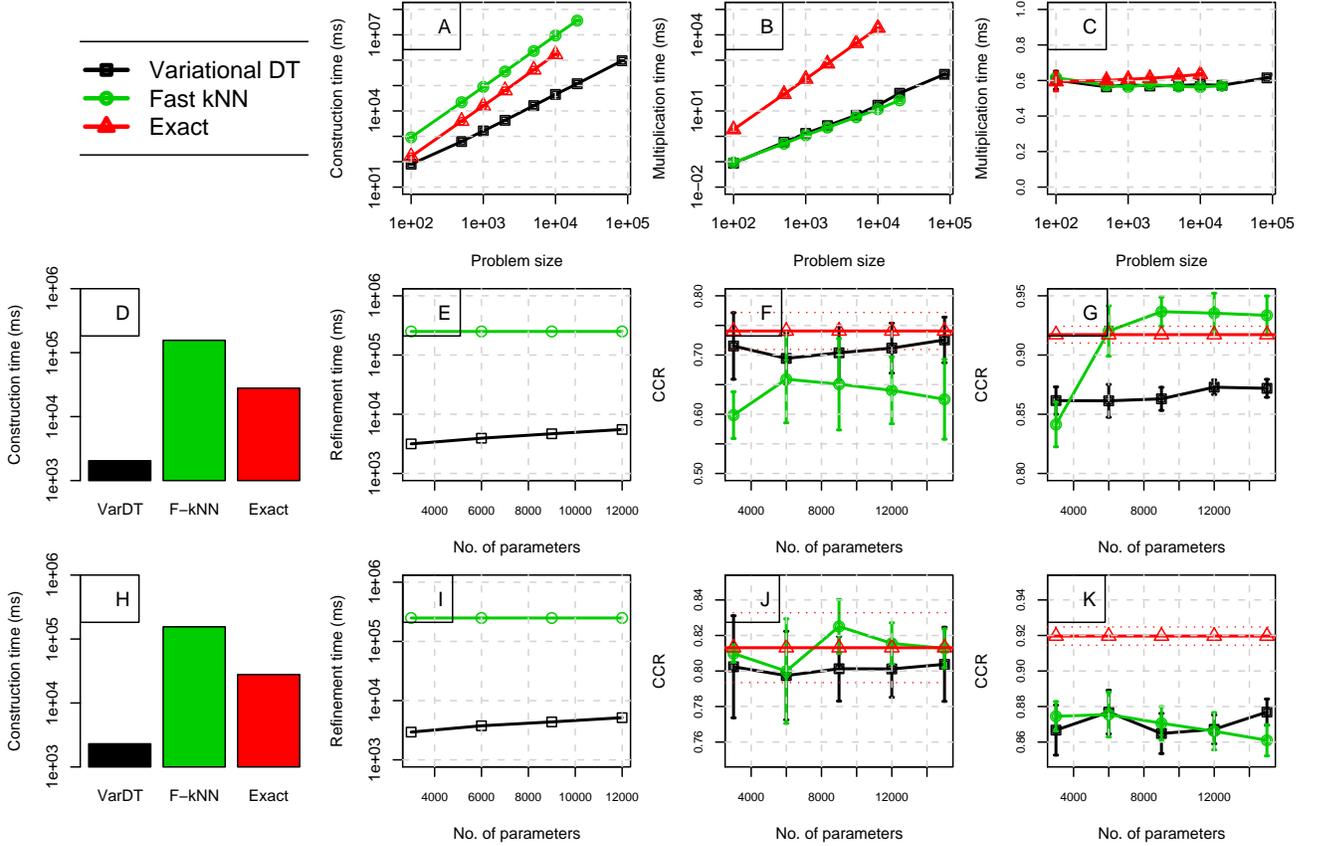

Figure 2: The experimental results for Variational DT, $k$NN and exact methods

In the second experiment, we study the efficiency and the effectiveness of the refinement process for the $k$NN and variationalDT models. To this end, first we build the coarsest $k$NN ($k = 2$) and VariationalDT ($|\mathcal{B}| = 2(N-1)$) models and then refine each model to higher levels of refinement. At each refinement level, we make sure both methods have the same number of parameters; that is, $|\mathcal{B}| = kN$. We stop refining both models when the number of parameters reaches $O(N \log N)$. The data sets used for this experiment are Digit1 and USPS from the set of SLL benchmark data sets (Chapelle et al., 2006). Both data sets consist of digit images; while Digit1 is an artificially generated data set, USPS contains images of handwritten digits. Each data set has 1500 examples, 241 features and 2 classes. The second (third) row in Figure 2 shows the results for Digit1 (USPS) data set. Figures 2D,H) show the construction times for initial coarse models. Again our method is much faster than the baselines in terms of construction time. In Figures 2E,I), we have measured the time needed to refine each model to the next level of refinement. As the figure illustrates, VariationalDT needs an order of magnitude less time for refinement than $k$NN. Moreover, at each refinement level, we measure the CCR for LP when the size of the labeled set is 10 (Figure 2F,J) and 100 (Figure 2G,K). The red flat line in both plots depicts the CCR for the exact model. As the plots show, $k$NN and VariationalDT behave differently for different data sets and different sizes of labeled data. While refinement improves the $k$NN's performance significantly in Figure 2G, it degrades the $k$NN's performance in Figure 2F,K. We attribute this abrupt behavior to the *uniform* refinement of $k$NN graph. More precisely, when the $k$NN graph is refined, the degrees of all nodes are uniformly increased by 1 regardless of how much the log-likelihood lower bound is improved. Our method, on the other hand, explicitly aims to increase the log-likelihood lower bound resulting in a non-uniform refinement of $P$ as well as more consistent behavior in terms of accuracy.

In the third experiment, we explore the applicability and the scalability of the proposed framework for very large data sets. The two data sets used in this experiment are taken from the Pascal Large-scale Learning Challenge.[1] The first data set, *alpha*, consists of 500,000 records with 500 dimensions. The second data set, *ocr*, is even larger with 3,500,000 records of 1156 features. Both data sets consist of 2 balanced classes. Table 2 shows the construction and propagations times as well as the number of model parameters when the Variational DT algorithm is applied. We could not apply other baseline methods for these data sets due to their infeasible construction times. Nevertheless, by extrapolating the graphs in Figure 2A, we guess the baselines would be roughly 3-4 orders of magnitude ($10^3 - 10^4$ times) slower. It should also be stressed that this experiment is specifically designed for showing the applicability of our framework for gigantic data sets and not necessarily showing its accuracy. However, even though we have not tuned the SSL parameters (like the labeling threshold), we still got $CCR = 0.56 \pm 0.01$ which is better than the random classifier. In conclusion, as a serial algorithm, our framework takes a reasonable time to construct and operate on these gigantic data sets. Furthermore, due to its tree-based structure, the Variational DT framework has the great potential to be parallelized which will make the algorithm even faster.

| Data set | $N$ | Param# | Const. | Prop. |
|---|---|---|---|---|
| **alpha** | 0.5 M | 1 M | 4.5 hrs | 11.7 min |
| **ocr** | 3.5 M | 7 M | 46.2 hrs | 93.3 min |

Table 2: Very large-scale results

# 6 Conclusions

In this paper, we proposed a very efficient approximation framework based on a variational dual-tree method to estimate, store and perform inference with the transition matrix for random walk on large-scale data graphs. We also developed an unsupervised optimization technique to find the bandwidth for the Gaussian similarity kernel used in building the transition matrix. Algorithmically, we extended the variational dual-tree framework with a fast multiplication algorithm used for random walk inference with applications in large-scale label propagation and eigendecomposition. We also provided a complexity comparison between our framework and the popular $k$-nearest-neighbor method designed to tackle the same large-scale problems. In experiments, we demonstrated that while both the $k$NN based method and our method do not compromise much in terms of accuracy compared to the exact method, our method outperforms the $k$NN based method in terms of construction time with order of magnitudes difference. We also showed that our framework scales to gigantic data sets with millions of records.

Our future directions for this work include adding an inductive feature to our framework to deal with new examples as well as exploring the theoretical aspects of the approximation made by the algorithm.

# Acknowledgements

This research work was supported by grants 1R01LM010019-01A1 and 1R01GM088224-01 from the NIH and by the Predoctoral Andrew Mellon Fellowship awarded to Saeed Amizadeh for the school year 2011-2012. The content of this paper is solely the responsibility of the authors and does not necessarily represent the official views of the NIH.

---

[1] http://largescale.ml.tu-berlin.de/